\pdfoutput=1

\documentclass[11pt]{article}

\usepackage[]{acl}

\usepackage{times}
\usepackage{latexsym}

\usepackage[T1]{fontenc}

\usepackage[utf8]{inputenc}

\usepackage{microtype}

\usepackage{graphicx}
\usepackage{multirow}
\usepackage{booktabs}
\usepackage{amssymb}
\usepackage{subfigure}
\usepackage{bbding}
\usepackage{pifont}
\usepackage{amsmath}
\usepackage{enumitem}

\usepackage{algorithm,algpseudocode}
\algnewcommand{\algorithmicforeach}{\textbf{for each}}
\algdef{SE}[FOR]{ForEach}{EndForEach}[1]
  {\algorithmicforeach\ #1\ \algorithmicdo}
  {\algorithmicend\ \algorithmicforeach}

\title{K-ESConv: Knowledge Injection for Emotional Support Dialogue Systems \\ via Prompt Learning}

\author{Wei Chen\textsuperscript{\rm 1}, Gang Zhao\textsuperscript{\rm 2}, Xiaojin Zhang\textsuperscript{\rm 1}, Xiang Bai\textsuperscript{\rm 1}, Xuanjing Huang\textsuperscript{\rm 2}, Zhongyu Wei\textsuperscript{\rm 2}\thanks{~~Corresponding Author.} \\ \textsuperscript{\rm 1}Huazhong University of Science and Technology, China \\
\textsuperscript{\rm 2}Fudan University, China \\ 
\{lemuria\_chen,xiaojinzhang,xbai\}@hust.edu.cn; \\ 21210980095@m.fudan.edu.cn; \{xjhuang,zywei\}@fudan.com}

\begin{document}
\maketitle
\begin{abstract}

Automatic psychological counseling requires mass of professional knowledge that can be found in online counseling forums. Motivated by this, we propose K-ESConv, a novel prompt learning based knowledge injection method for emotional support dialogue system, transferring forum knowledge to response generation. We evaluate our model on an emotional support dataset ESConv, where the model retrieves and incorporates knowledge from external professional emotional Q\&A forum. Experiment results show that the proposed method outperforms existing baselines on both automatic evaluation and human evaluation, which shows that our approach significantly improves the correlation and diversity of responses and provides more comfort and better suggestion for the seeker.

\end{abstract}

\section{Introduction}
\label{section:intro}
Psychologists provide consultation to comfort seekers via face-to-face conversation or online forum counseling, helping them relieve negative emotion and solve the problems they meet in daily life \cite{burleson2003emotional,reblin2008social}. Some classical forums include mentalhealthforum\footnote{https://www.mentalhealthforum.net/}, Breakthrough\footnote{https://Breakthrough.com/} and TalkTherapy section\footnote{https://www.reddit.com/r/TalkTherapy/} in Reddit.


In consideration of high costs of face-to-face psychological counselling, automatic counseling pattern has received lots of interests recently  \cite{liu2021towards}. Although large-scale pre-trained language models (PLMs) \cite{radford2018improving,radford2019language,zhang2019dialogpt,roller2020recipes,chen2022dialogved,chen2022contextual} have shown remarkable performance on a range of dialogue response generation tasks, existing PLMs-driven psychological consultation systems are limited in two points. (1) The scarcity of psychological counselling data makes it difficult to fully train the model. (2) Distinguished from general dialog systems, psychological consultation needs more professional knowledge. 

\begin{figure}
\centering
\includegraphics[width=1.0\columnwidth]{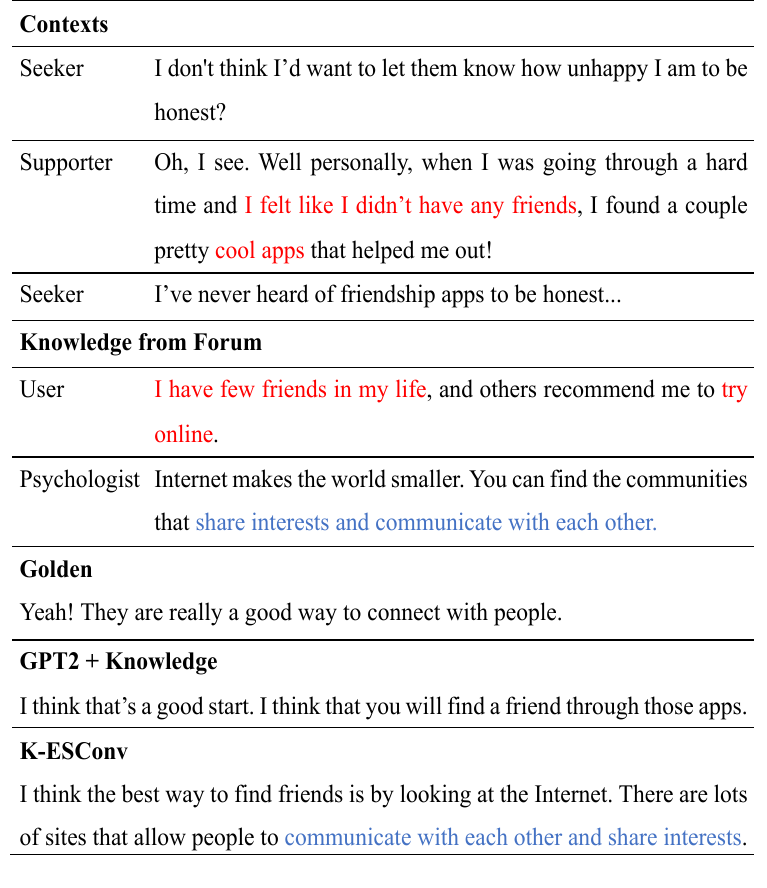}
\caption{Sample conversation on emotional support. Compared with GPT2 with knowledge injection, the response generated by K-ESConv reflects "friendship apps" in the context and makes some extensions.}
\label{fig:intro}
\end{figure}

An example is shown in Figure~\ref{fig:intro}. Through our observations, professional knowledge required in psychological counselling can mostly be found in online counseling forums. Therefore, we consider employing forums as external knowledge source, advancing the promotion of automatic psychological counselling. For better fusing knowledge, we propose K-ESConv, a novel knowledge injection method via prompt-learning to generate context-aware and knowledge-aware responses for emotional support dialogue system. We conduct experiments on ESConv dataset \cite{liu2021towards} and choose PsyQA \cite{sun2021psyqa} as an external knowledge base. Knowledge selection is implemented by a pretrained dense retriever \cite{karpukhin2020dense}. We compare K-ESConv with both knowledge-free and knowledge-grounded baseline models, and the preliminary experimental results show that K-ESConv has a better performance on diversity and professionalism for response generation.

\section{Related Work}
\label{section:related-work}




\paragraph{Knowledge Grounded Response Generation} ~ {Benefiting from knowledge-equipped dialogue datasets \cite{zhou2018dataset,dinan2018wizard}, knowledge-based dialogue systems have shown great effects in improving the informativeness and diversity of responses \cite{zhao2020knowledge,shuster2021retrieval,komeili2021internet,xu2021retrieval}. Most works focus more on knowledge selection. For example, DIALKI \cite{wu2021dialki} leverages the document structure to provide dialogue-contextualized passage encodings and better locate knowledge relevant to the conversation.}

 
\paragraph{Prompt Learning for Response Generation}{Prompt learning is to fine-tune a lightweight trainable module to adapt PLM to downstream specific tasks, such as dialogue response generation. For example, Prefix-Tuning \cite{li2021prefix} considers tuning prompts with textual prefix, where some task-specific "soft tokens" are prepended to the source text and only the prompt token encoding module are trainable. Considering dialogue responses are affected by contextual information, DialogPrompt \cite{gu2021response} is proposed to learn a prompt encoder that encourages context-aware prompt learning.}

\section{Method}
\label{section:method}

\begin{figure*}
\centering
\scalebox{0.43}{
\includegraphics{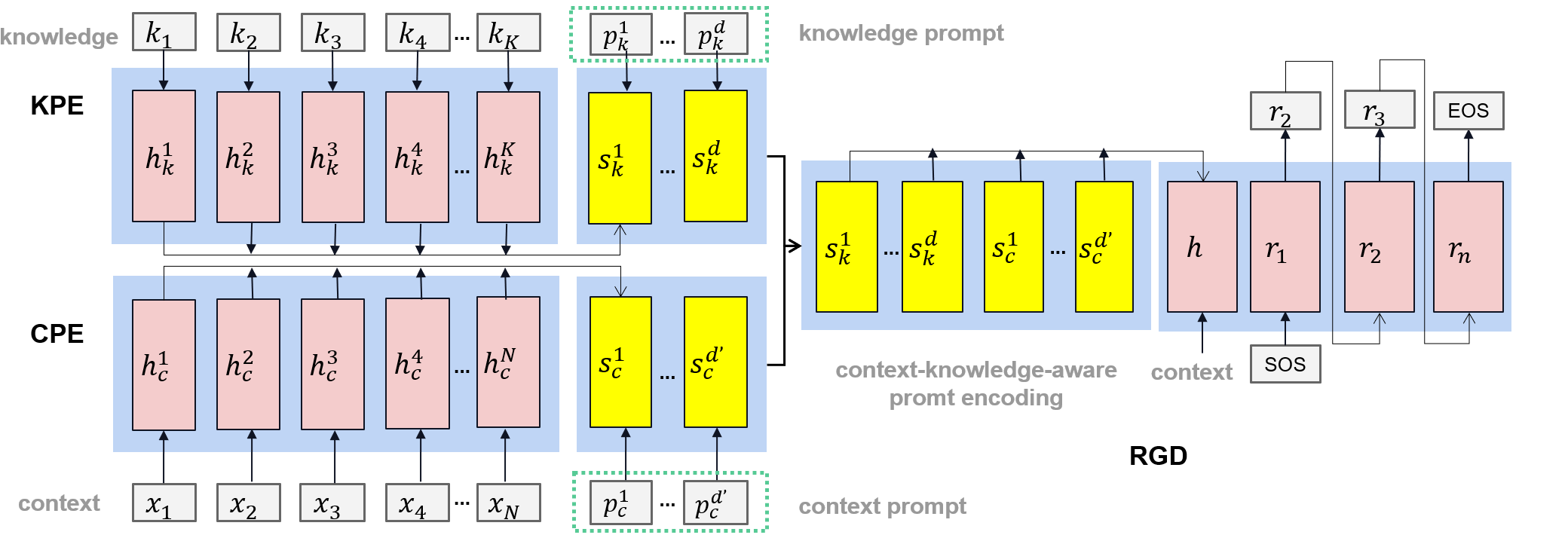}}
\caption{K-ESConv composed of context-aware prompt encoder (CPE), knowledge-aware prompt encoder (KPE) and response generation decoder (RGD). The blue block denotes PLM for prompt encoding or response generation. The pink block denotes the hidden states in PLM. The yellow block denotes the prompt encoding that encourages RGD to elicit relevant and diverse responses.}
\label{fig:my_label}
\end{figure*}

\subsection{Task Formalization}

Different from traditional knowledge-grounded response generation tasks, there is no ground truth knowledge document for a given context in our task,  the knowledge base we use in this paper is an emotional support dataset consisting of question-answer (QA) pairs, which we assume that can implicitly provide specialized knowledge and patterns for emotional support conversations. Formally, given a context $\mathcal{U}=\{x_i\}_{i=1}^N$ consisting of a sequence of $N$ tokens, and a QA knowledge base $\{(Q_i, A_i)\}_{i=1}^M$  consisting of $M$ samples, the goal of our task is to select a knowledge from knowledge base for the context and generate the corresponding response $\mathcal{R}$.

\subsection{Knowledge Selection}

Due to the lack of ground truth knowledge, we adopt DPR \cite{karpukhin2020dense}, a pre-trained retriever for open QA, as our knowledge selector. For a given context, we retrieve the Top-1 most similar questions by their inner-products of dense embeddings, and the corresponding answers are regarded as the factual knowledge.

\subsection{Prompt learning for context-aware and knowledge-aware adaptation}
Our model framework is shown in Figure \ref{fig:my_label}. In our approach, prompt learning is adapted for context-aware and knowledge-aware purpose. And the prompt encoding is optimized to steer the PLM to context and knowledge. 

Specifically, K-ESConv consists of a knowledge-aware prompt encoder (KPE), a context-aware prompt encoder (CPE) and a response generation decoder (RGD), all based on pretrained language models (PLMs). KPE and CPE takes context and retrieved knowledge as input respectively, and obtain the hidden vectors of virtual knowledge-aware prompts $\{p_k^i\}_{i=1}^d$ and context-aware prompts $\{p_c^i\}_{i=1}^{d'}$. Subsequently, the prompt encodings are reparameterized by a MLP layer in order to improve the stability of optimization :
\begin{align}
    s_k^i = MLP(KPE(k_1,k_2,\cdots,k_K,p_{k}^{<i}))\\
    s_c^i = MLP(CPE(x_1,x_2,\cdots,x_N,p_c^{<i}))
\end{align}

\subsection{Prompt Combination for Response Generation}
Finally, the concatenation of context-aware prompt encoding and knowledge-aware prompt encoding $[s_k^1,s_k^2,\cdots,s_k^d,s_c^1,s_c^2,\cdots,s_c^{d'}]$ are taken as past hidden states of RGD. Then the context $\{x_i\}_{i=1}^N$ is fed to RGD for response decoding. Our optimization objective is to minimize the generative loss of RGD.

In retrospect to our model, there are totally three PLMs, in which CPE and KPE are trainable, but RGD is frozen. In training phase, the generative loss is minimized to supervise CPE and KPE to generate context-aware prompt encoding and knowledge-aware prompt encoding respectively.



\section{Experiments}
\label{section:experiment}

\subsection{Dataset}
We evaluate our model on ESConv and choose PsyQA as the external database. Emotional Support Conversation (ESConv), is a manually labeled multi-turn dialogue dataset that contains conversations about emotion comfort in English. In ESConv, the mean length of response is 16. PsyQA is a chinese dataset of mental health support in the form of Question and Answer (QA). Each QA pair includes one question from the psychological comfort seeker and one answer from help provider. In PsyQA, the mean length of Question and Answer is 114 and 36 respectively.Baselines, Implementation details and evaluation metrics are in Appendix \ref{baselines}, Appendix \ref{Implementation Details}.
To evaluate the impact of context-aware prompt encoding and knowledge-aware prompt encoding, we compare the performance of all baselines and K-ESConv. 

\subsection{Evaluation Metrics}
\paragraph{Automatic Evaluation}~{We report automatic metrics for evaluating NLG including BLEU-1/2~\cite{papineni2002bleu}, DIST-1/2/3~\cite{https://doi.org/10.48550/arxiv.1510.03055} and ROUGE-L~\cite{lin2004rouge}. Additionally, we report the BERTScore (BRTSc)~\cite{bert-score} of the generated response and corresponding retrieved knowledge to measure the performance of knowledge injection.}


\paragraph{Human Evaluations}~{We randomly sample 200 dialogues from the test set and generate responses using our model and two baselines. Then 2 annotators with relevant backgrounds are prompted to choose the better response based on indicators in \cite{https://doi.org/10.48550/arxiv.2106.01144}}: (1) Fluency:
which one is more fluent? (2) Identification: which one is more
helpful in identifying the seeker’s problems? (3) Comforting: which
one is more skillful in comforting the seeker? (4) Suggestion: which
one provides more helpful suggestions? (5) Overall: generally, which emotional support do you prefer?

\begin{table*}[]
\centering
\scalebox{0.8}{
\begin{tabular}{llcccccc}
\toprule
\multirow{1}{*}{\textbf{Type}} & \multirow{1}{*}{\textbf{Model}} & \multirow{1}{*}{\textbf{BLEU-1}} & \multirow{1}{*}{\textbf{DIST-1}} &
\multirow{1}{*}{\textbf{DIST-2}} & \multirow{1}{*}{\textbf{DIST-3}} & 
\multirow{1}{*}{\textbf{ROUGE-L}} & \multirow{1}{*}{\textbf{BRTSc}}  \\ \toprule
\multirow{7}{*}{\textbf{w/o. knowledge}} & Transformer \cite{vaswani2017attention} & 17.25 & 1.25 & 7.29 & - & 14.68 & - \\
 &        MoEL \cite{lin2019moel} & 16.02 & 2.71 & 14.92 & - & 14.26 & - \\
 &        MIME \cite{majumder2020mime} & 16.15 & 2.56 & 12.33 & - & 14.83 & - \\
 &       GPT-2 \cite{radford2019language} & 16.96 & 3.04 & 18.80 & 41.57 & 15.02 & - \\
 &    DialoGPT \cite{zhang2019dialogpt} & 17.41 & 3.03 & 19.26 & 42.19 & 14.82 & - \\
 &  DialogPrompt \cite{gu2021response} & 17.37 & 4.11 & 24.37 & 52.60 & 14.29 & - \\
 &  MISC \cite{https://doi.org/10.48550/arxiv.2203.13560} & 17.71 & 3.65 & 22.25 & - & 15.57 & -  \\
 \midrule
\multirow{4}{*}{\textbf{w. knowledge}} & GPT2 + Knowledge & 17.82 & 3.22 & 19.69 & 42.22 & 15.13 & 83.4 \\
 &  DialoGPT + Knowledge  & 17.45 & 3.13 & 19.44 & 42.66 & 15.51 & 83.9 \\
 & Prefix-Tuning \cite{li2021prefix}  &  17.65   & 3.71     & 22.60      & 49.03      &  14.01     & 83.8 \\
 & PoKE \cite{https://doi.org/10.48550/arxiv.2210.12640} & 18.41 & 3.73 & 22.03 & - & \textbf{15.84} & - \\
 \midrule
\textbf{Ours} & \textbf{K-ESConv} & \textbf{18.83} & \textbf{4.37} & \textbf{25.46} & \textbf{54.34} & 15.37 & \textbf{84.9} \\ \bottomrule
\end{tabular}}\caption{Automatic evaluations metrics on ESConv. In K-ESConv, we set context-aware prompt size as 10 and knowledge-aware prompt size as 5.}
\label{tab:experiment-result}
\end{table*}

\begin{table}[]
\centering
\scalebox{0.8}{
\begin{tabular}{lcccc}
\toprule
\multirow{2}{*}{\textbf{Ours vs}} 
 & \multicolumn{2}{c}{\textbf{w.knowl}} & \multicolumn{2}{c}{\textbf{w/o.knowl}} \\
 & \textbf{Win} & \textbf{Lose} & \textbf{Win} & \textbf{Lose} \\
\toprule
\textbf{Fluency}        & \textbf{50.0}  & 31.5 & \textbf{63.2} & 24.8 \\
\textbf{Identification} & \textbf{52.6}  & 34.2 & \textbf{75.6} & 23.2 \\
\textbf{Comforting}     & \textbf{57.9}  & 36.8 & \textbf{80.4} & 15.7 \\
\textbf{Suggestion}     & \textbf{55.2}  & 28.9 & \textbf{73.2} & 16.8 \\
\textbf{Overall}        & \textbf{57.5}  & 37.2 & \textbf{82.0} &
15.5\\
\bottomrule
\end{tabular}}
\caption{Human evaluations results. DialoGPT represents for knowledge-free baselines and DialoGPT with retrieved knowledge represents for knowledge-grounded baselines. Kappa score \cite{doi:10.1177/001316447303300309} reaches 0.512, indicating a moderate level of agreements.}
\label{tab:human-evaluation}
\end{table}

\subsection{Main Results}
To evaluate the effectiveness of our model, we conduct experiments on knowledge-free baselines and knowledge-grounded baselines. Each category covers the corresponding SOTA model i.e. MISC\cite{https://doi.org/10.48550/arxiv.2203.13560} in knowledge-free baselines and POKE\cite{https://doi.org/10.48550/arxiv.2210.12640} in knowledge-grounded baselines. Details are in Appendix \ref{baselines}.

The overall result of the comparison is shown in Table \ref{tab:experiment-result}. We have following findings.
\begin{itemize}[leftmargin=*]
\item[$\bullet$] The performance of most knowledge-grounded baselines is better than that of knowledge-free baselines, especially on aspects of DIST. It indicates that knowledge injection helps generate more diverse responses. Furthermore, it proves the effectiveness of professional knowledge in psychological counseling.
\item[$\bullet$] The performance of K-ESConv is better than that of knowledge-grounded baselines, slightly in BLEU and significantly in DIST. It indicates that the response in our method is superior in terms of correlation and diversity. Moreover, it proves that KPE promotes the diversity of response.
\item[$\bullet$] The performance of K-ESConv is slightly better than that of knowledge-grounded baselines in BERTScore. It indicates that the response in our method condenses more information from retrieved external knowledge. Furthermore, it proves that our approach could generate more knowledgeable and informative response.
\end{itemize}

The human evaluation results are shown in Table \ref{tab:human-evaluation}. We have following findings.
\begin{itemize}[leftmargin=*]
		\item[$\bullet$] The performance of K-ESConv is much better than that of knowledge-free baselines on all indicators, especially on aspects of Identification and Comforting. It proves the significant effect of knowledge injection on psychological counseling.
		\item[$\bullet$] The performance of K-ESConv is better than that of knowledge-grounded baselines, mainly on aspects of Identification, Comforting and Suggestion. It indicates that our generation could reflect more information related to the comfort seeker and offer more helpful comfort based on it. Furthermore, experiment results prove that our method could generate more knowledgeable and professional responses.
\end{itemize}
\subsection{Case Study}
Figure \ref{fig:intro} shows one case in the psychological counseling scenario. In Contexts, the comfort seeker expresses his anxiety on lack of friends, and mentioned the "online friendship apps". In counseling forum, psychologists have solved similar problems and elaborates the effect of Internet on making new friends. Employing the case as external knowledge, K-ESConv makes detailed suggestions on expanding online circle of friends and "cites" the advantages of making friends online in knowledge from forum. 
\par Due to page limitations, we put the analysis of the effect of prompt size in appendix \ref{Effect of Prompt Size}.

\section{Conclusion}
\label{section:conclusion}
In this paper, we propose K-ESConv, a novel knowledge injection method for ESConv task via prompt learning, transferring forum knowledge to response generation. Evaluation on ESConv proves the superiority of our model in the correlation and diversity of responses. We expect to explore the validity of our K-ESConv on general dialogue benchmarks in the future. 

\section{Limitation}
\label{section:limitation}
It is possible that multiple pieces of knowledge correspond to the contexts. Therefore, how to select multiple knowledge pieces and inject them into response generation remains to be explored.   
\bibliography{anthology,custom}
\bibliographystyle{acl_natbib}

\appendix

\section{Baseline}
\label{baselines}
To evaluate the effectiveness of our model, we conduct experiments on many baselines. We sort all baselines as two categories:
\paragraph{Knowledge-Free Baselines} In this section, we introduce knowledge-free baseline models. (1) \textbf{Transformer} \cite{vaswani2017attention}, the standard Transformer model. (2)     \textbf{MoEL} \cite{lin2019moel}, a Transformer-based model which softly combines the response representations from different transformer decoders. (3) \textbf{MIME} \cite{majumder2020mime}, another extension of Transformer model which mimics the emotion of the speaker. (4) \textbf{GPT2}\cite{radford2019language}, the unidirectional pretrained language model (5) \textbf{DialoGPT} \cite{zhang2019dialogpt}, which is a GPT-based model pretrained on large-scale dialog corpora. (6) \textbf{DialogPrompt} \cite{gu2021response}, a prompt learning method for general dialogue response generation, where prompts are dynamically encoded for context-aware adaptation. (7) \textbf{MISC} \cite{https://doi.org/10.48550/arxiv.2203.13560}, infers the user’s fine-grained emotional status using COMET\cite{https://doi.org/10.48550/arxiv.1906.05317}, and responds skillfully by a mixture of strategy.

\paragraph{Knowledge-Grounded Baselines} In this section, we introduce baselines with knowledge injection, where knowledge and context are concatenated and are fed to model. (1) \textbf{GPT2 + Knowledge}, GPT2 model with knowledge injection. (2) \textbf{DialoGPT + Knowledge}, DialoGPT model with knowledge injection. (3) \textbf{Prefix-Tuning + Knowledge} \cite{li2021prefix}, a prompt learning based model with knowledge injection. (4) \textbf{POKE}\cite{https://doi.org/10.48550/arxiv.2210.12640}, a prior knowledge enhanced model with latent variable.

\section{Implementation Details}
\label{Implementation Details}
We used GPT-2\cite{radford2019language} as the backbone pretrained language model for our model. Our implementation was based on the Huggingface Transformer Repository\cite{https://doi.org/10.48550/arxiv.1910.03771}. We set prompt sequence length as 10 and 5 for context-aware prompt encoder and knowledge-aware prompt encoder respectively. The batch size is set to 8. In the generation phase, we used the top-1 sampling for response decoding. Our model is optimized using the AdamW\cite{https://doi.org/10.48550/arxiv.1711.05101} whose initial learning rate is 5e-5. We used a linear learning rate scheduler with 200 warm-up steps. We train our model on one GPU of Nvidia 3090.

\begin{figure}[!ht]
\centering
\includegraphics[width=1.0\columnwidth]{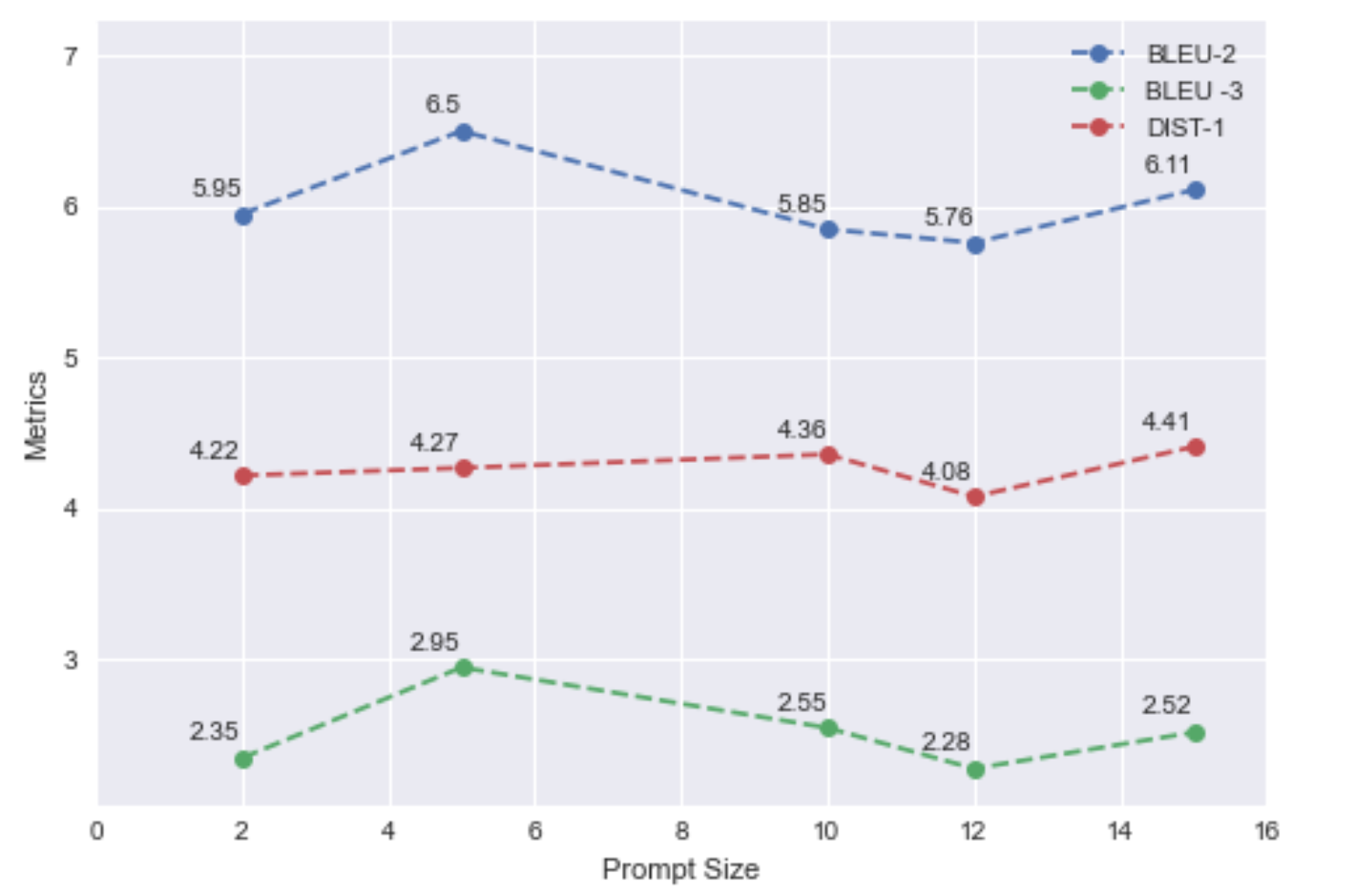}
\caption{Effects of knowledge prompt size on the performance when context prompt size is fixed to 10}
\label{fig:knowledge_prompt_size}
\end{figure}

\begin{figure}[!ht]
\centering
\includegraphics[width=1.0\columnwidth]{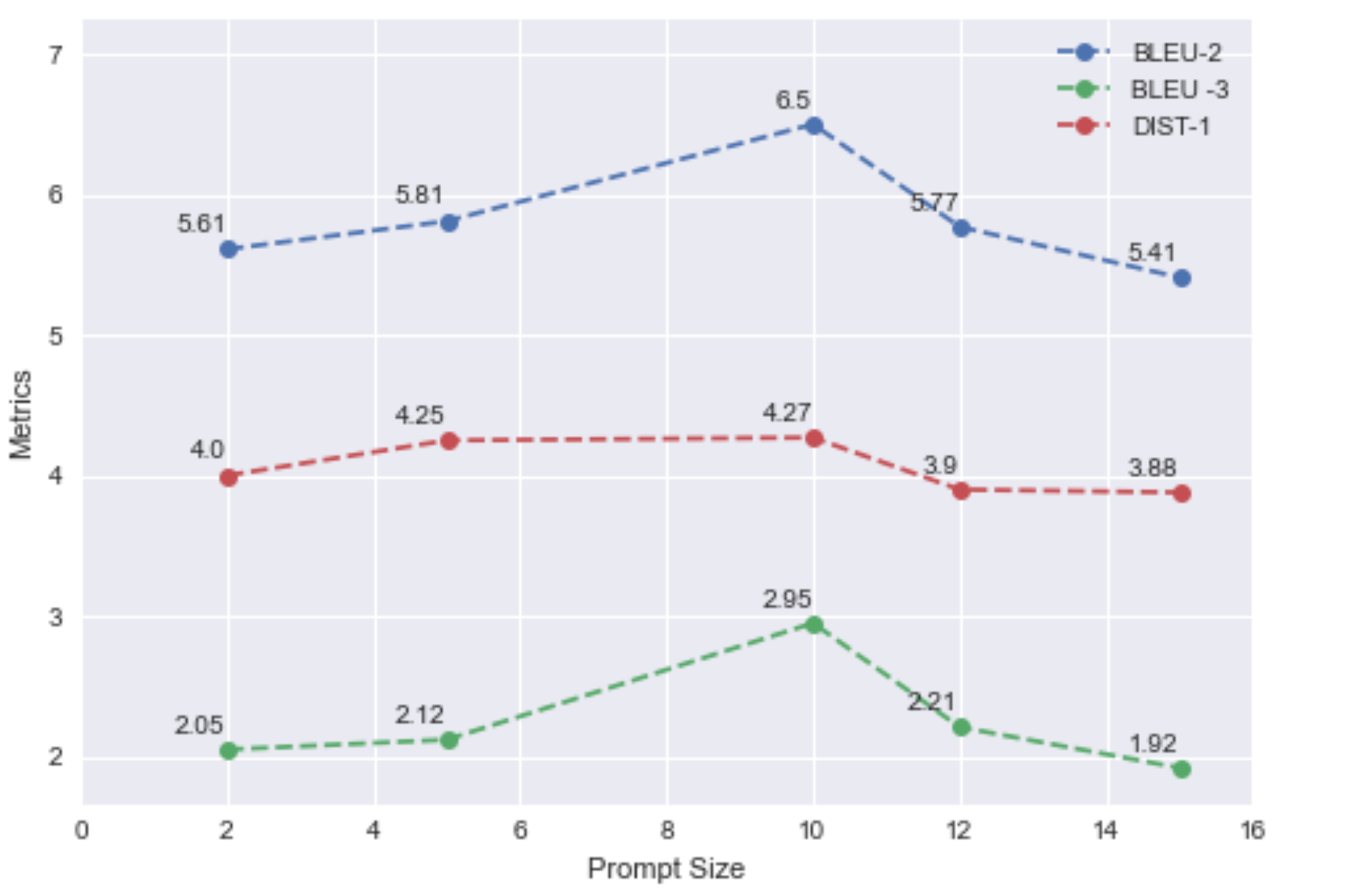}
\caption{Effects of context prompt size on the performance when knowledge prompt size is fixed to 5}
\label{fig:context_prompt_size}
\end{figure}

\section{Effect of Prompt Size}
\label{Effect of Prompt Size}
One of the most important hyperparameters in our approach is the length of prompts that affects the degree of PLMs being prompted. Thus we conduct experiments on the impact of the prompt size by tuning the knowledge prompt size and context prompt size in K-ESConv. The results in Figure \ref{fig:knowledge_prompt_size} and Figure \ref{fig:context_prompt_size} shows that the difference of both prompt size leads to slight change in all metrics, showing that the performance of K-ESConv is not sensitive to the prompt size. The performance in BLEU reaches top when the knowledge-aware prompt size are set to 5. When the knowledge-aware prompt size is set to 10, the DIST of the generation reaches top, which means that PLM elicits responses with highest diversity and informativeness. Besides, when context prompt size is set to 10, BLEU and DIST both reach top. In order to balance the computation efficiency and evaluation performance, we select 10 and 5 as optimal choices for context-aware prompt size and knowledge-aware prompt size. 

\end{document}